\documentclass[10pt,doublespacing]{elsart}
\usepackage{latexsym,amssymb,amsmath,graphicx,epsfig,cite,bbm,float,epsf,graphics}
\journal{Digital Signal Processing}

\renewcommand{\vec}[1]{\mbox{\boldmath${#1}$}}

\newcommand{\ei}{\end{itemize}}
\newcommand{\bi}{\begin{itemize}}

\newcommand{\MB}{\left[\begin{array}}
\newcommand{\ME}{\end{array}\right]}

\makeatletter

\newcommand{\Rmnum}[1]{\expandafter\@slowromancap\romannumeral #1@}
\makeatother

\begin{document}
\begin{frontmatter}

\title{A Novel Training Algorithm for HMMs with Partial and Noisy Access to the States}
\vspace{-0.5in}
\author[1]{Huseyin Ozkan},
\ead{huozkan@ku.edu.tr}
\author[2]{Arda Akman},
\ead{arda.akman@turktelekom.com.tr}
\author[1]{Suleyman S. Kozat\corauthref{cor}}
\corauth[cor]{Corresponding author.}
\ead{skozat@ku.edu.tr}
\address[1]{Department of Electrical and Computer Engineering, Koc University, Istanbul, Tel: 90 212 3501840.}
\address[2]{Turk Telekom Group R$\&$D, Ankara, Tel: 90 312 5556700.}
\vspace{-0.3in}
\begin{abstract}
This paper proposes a new estimation algorithm for the parameters of an HMM as to best account for the observed data. In this model, in addition to the observation sequence, we have \emph{partial} and \emph{noisy} access to the hidden state sequence as side information. This access can be seen as ``partial labeling" of the hidden states. Furthermore, we model possible mislabeling in the side information in a joint framework and derive the corresponding EM updates accordingly. In our simulations, we observe that using this side information, we considerably improve the state recognition performance, up to $70\%$, with respect to the ``achievable margin" defined by the baseline algorithms. Moreover, our algorithm is shown to be robust to the training conditions.
\end{abstract}
\begin{keyword}
HMM training, ML estimator, side information, partial, noisy.
\end{keyword}
\end{frontmatter}
\section{Introduction}
\begin{figure}[t]
\centerline{\epsfxsize=7cm \epsfbox{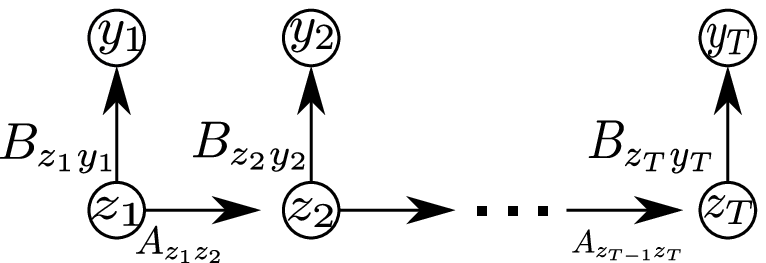}
\hspace{0.3in} \epsfxsize=9cm \epsfbox{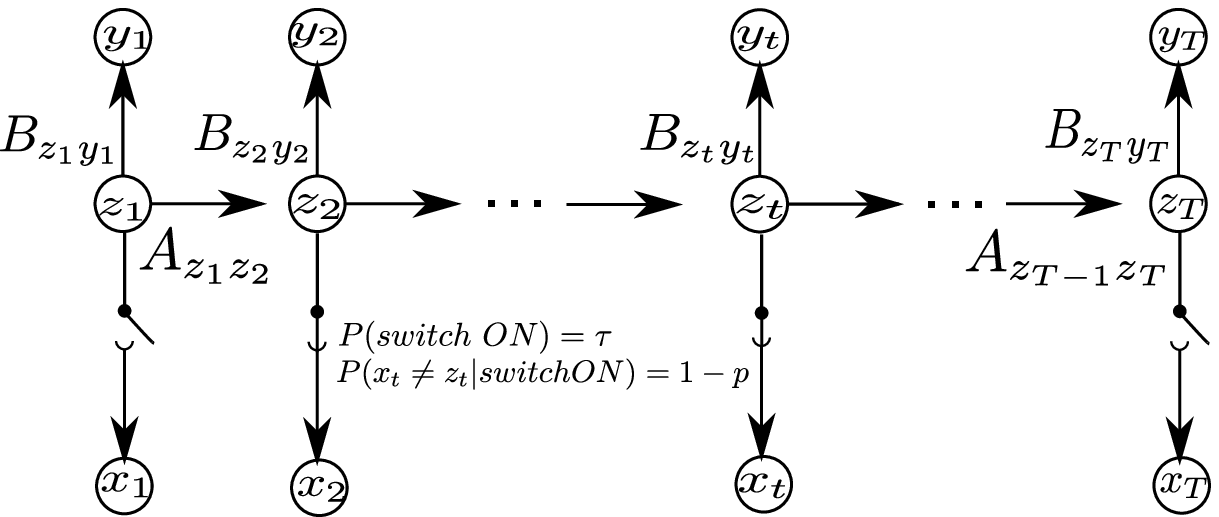}}
\centerline{\hspace{1.6in}(a)\hspace{3.2in}(b)\hfill}
\caption{(a) An example HMM with discrete-time  finite-state, $z_t$, and observations, $y_t$, of a finite alphabet. (b) Observation Model with Noisy and Partial Access to the State Sequence.}
\label{HMMfigure}
\end{figure}

In a wide variety of applications in time series analysis ranging from speech processing \cite{HMM,DREFH,SRUGHSMM,APIOAHMMWDMFSR,  APPT, LSDTHMMFSR, ELLAFSR}, bioinformatics \cite{HMMIBSA, HMMbi}
to natural language processing \cite{SIFPHMM, TETWAPM,DBWRHT,LHMMSFIE}, the observation sequence is represented as a stochastic process, depending on another stochastic process which generates a sequence of hidden (unobserved) states. With certain properties regarding the observations as well as
the states, this is known as Hidden Markov Model (HMM) \cite{HMM}. In this paper, we particularly concentrate on discrete-time finite-state HMM with finite alphabet, which is described by two random variables: the hidden state $z_t$ and the observation $y_t$. The state sequence forms a stochastic, discrete-time Markov chain and the probability of an observation $y_t$ does only depend on the state $z_t$. Hence, as shown in Fig. \ref{HMMfigure}a, an HMM is completely characterized by
the state transition probabilities, $A_{ij}$, the
observation emission probabilities, $B_{ij}$, and the initial state
probabilities $\pi_i$. A detailed description of the model can be found in \cite{HMM}. Estimation of these model parameters, $A_{ij}$, $B_{ij}$ and $\pi_i$, is an important problem in applications using HMM \cite{HMM, APPT, TETWAPM, DBWRHT, HMMIBSA, SIFPHMM,LHMMSFIE,ALOPHMM, LSDTHMMFSR, HMMbi}. Since there is no closed form solution for the set of parameters that maximizes the probability of the
observation sequence given the model, instead, iterative algorithms such as the Expectation-Maximization (EM) algorithm \cite{EMalgorithm} (or equivalently the Baum-Welch method \cite{BaumWelch}) is used to obtain
a local optimal solution \cite{HMM}. In this paper, we derive a new set of iterative EM equations that yield a locally optimal solution for the model parameters, when the ordinary model of the observation sequence, e.g., as in \cite{HMM}, is different. In our model, in addition to the observation sequence $y_t$, we observe a part of the hidden state sequence as side information. More precisely, at every time instant $t$, we observe the hidden state with probability $\tau$, i.e., with $1-\tau$ probability the state is hidden. This gives partial access to the state sequence and hence, leads to a new model different from the ordinary HMM. We emphasize that the state observations are not necessarily confined to a time interval but may even be sparsely and randomly distributed along the complete time span of the application. In the limiting case, if $\tau$ is 0, then there would be no state observation, and we recover the ordinary, unsupervised HMM training. Therefore, our model provides a generalized framework by letting partial access to the state sequence. Moreover, we also allow that a state observation might be corrupted with noise such that if $z_t$ is ever observed, say as $x_t$, then $P(z_t \neq x_t) = 1-p$, as shown in Fig. \ref{HMMfigure}b. Under these new circumstances, we explicitly provide the mathematical derivations of the new set of iterative EM equations that incorporates the side information and estimate the model parameters accordingly. In these derivations, the probability that a state observation is incorrect, $1-p$, is assumed to be known and it is provided to our algorithm as a parameter, $p$, which defines the confidence on the side information. Simulations show that our method is robust to the confidence parameter $p$, even if it does not exactly match with the underlying true quality of the side information, $p_{\mathrm{true}}$.

Since the hidden state sequence is partially observed,
our work falls into the category of Partially Hidden Markov Model (PHMM) training (note that this
term is used in \cite{PHMM} in a different context). Similar to semi-supervised learning,
PHMMs use both ``labeled'' (in our context the state information) and ``unlabeled'' data to obtain improved model training. Such an approach is suitable, when we have access to a limited amount of labeled data along with a large amount of unlabeled data. This happens, as an example, in speech processing applications \cite{ELLAFSR}, where labeling, i.e., transcription, is naturally costly \cite{ELLAFSR,SSL}, hence only limited amount is affordable, and transcriptions may contain errors. Furthermore, by allowing noisy access to the states, we model ``mislabeling'' event that may occur during labeling stage. PHMMs, to the best of our knowledge, date back to the studies \cite{TETWAPM,DBWRHT,LHMMSFIE} in the area of Natural Language Processing. In these studies, tagged text, corresponding to the known states of a PHMM, are first analyzed through a relative frequency modeling to construct an initial model, then this model is fed into the ordinary HMM training algorithm. However, these studies do not rigorously show how the partial state information is incorporated within the ordinary HMM parameter learning framework. The Maximum Likelihood Estimator (MLE) for the model parameters in a special case
of PHMMs, where only a certain state from the state space in the underlying Markov chain is known, is theoretically (consistency and asymptotic normality of the estimator) analyzed in \cite{SIFPHMM}. However, the equations for computing the MLE (using the EM algorithm or other Likelihood maximization techniques) in this special case of PHMM is not derived. In \cite{ALOPHMM}, iterative EM equations for a general case, where each observation can only belong to a pre-defined set of acceptable states are given, but no complete derivation is provided. On the contrary, we explicitly derive the new set of iterative EM equations for the PHMM parameter learning problem, when there is partial access to the underlying hidden state sequence.
Furthermore, the partial observation of the state sequence might be prune to noise in our model and this case is not considered in the existing literature.

After we provide the brief description of the basic HMM framework and the parameter estimation equations in Section II,
we derive the new set of iterative EM equations that incorporates partial and noisy access to the state sequence
as side information in Section III. Simulations are presented in Section IV and the paper concludes with
final remarks in Section V.
\section{Problem Description}
\label{sec:Problem Description}
In this section, we briefly describe the basic HMM framework \cite{HMM}. For the sake of notational simplicity, we study discrete-time finite-state HMM with finite alphabet. However, our derivations for incorporating the side information in Section III can be readily extended to the case, where the observations come from a continuous distribution and outcomes are vectors. A discrete-time HMM with finite alphabet is formally a Markov model, for which
we have a sequence of observations, $y_t$, drawn from a finite alphabet $V=\{v_1,v_2,...,v_{N_v}\}$, i.e.,
$y_t \in V, 1\leq t\leq T$. We also have a sequence of hidden (unobserved) states $z_t \in S = \{s_1,s_2,...,s_{N_s}\}$, where $S$ is the set of possible states, generated from a Markov process, i.e., $P(z_t|z_{t-1},z_{t-2},...,z_1)=P(z_t|z_{t-1})$. The observation sequence, $y_t$, is generated based on the state sequence $z_t$, i.e., $P(y_t|z_t,z_{t-1},...,z_{1},y_{t-1},...,y_{1}) = P(y_t|z_t)$. We consider $A$ as the transition matrix, where $A_{ij}$ represents the transition probability from state $s_i$ to $s_j$, $A_{ij}=P(z_t=s_j|z_{t-1}=s_i)$. Similarly, $B$ is the observation probabilities at each state, i.e., $B_{ij}=P(y_t=v_j|z_t=s_i)$. In order to complete the HMM observation model, we also define the initial state distribution vector as $\pi_i=P(z_1=s_i)$. Thus, $\lambda=(A,B,\pi)$ represents the parameter set that completely characterizes the HMM model shown in Fig. \ref{HMMfigure}a.

When we have access to only the observation sequence, i.e., without labels, then the iterative EM equations, which provide a locally optimal solution for the HMM parameters $\lambda$, are given in \cite{HMM}. These equations are obtained through the likelihood maximization, i.e., $\arg\max_{\lambda}P(Y|\lambda)$, $Y=\{y_1,y_2,...,y_T\}$, which is carried out with the well known forward-backward procedure \cite{FBP1, FBP2}. To describe this procedure, we first define the forward variable, $\alpha_t(i)$, along with the recursion in \cite{HMM} as
\begin{align}
\alpha_t(i)   &= P(y_1,y_2,...,y_t,z_t=s_i|\lambda) \nonumber \\
              &= B_{iy_{t}}\sum_{j=1}^{N_s}\alpha_{t-1}(j)A_{ji}, \text{  }  \alpha_1(i) = \pi_iB_{iy_1}, \text{  } 2\leq t\leq T, \label{eq:recalpha}
\end{align}
which is the probability of observing $Y_{1}^{t}=\{y_1,y_2,...,y_t\}$ and being at state $z_t=s_i$, given the model $\lambda$.
Similarly, the backward variable is given by
\begin{align}
\beta_t(i)  &= P(y_{t+1},y_{t+2},...,y_T|z_t=s_i,\lambda) \nonumber \\
            &= \sum_{j=1}^{N_s}\beta_{t+1}(j)A_{ij} B_{jy_{t+1}}, 1\leq t\leq T-1,  \beta_T(i) = 1, \label{eq:recbeta}
\end{align}
which is the probability of observing $Y_{t+1}^{T}=\{y_{t+1},y_{t+2},...,y_T\}$, given the state $z_t=s_i$ and
the model. Based on these, we define the probability of transition at time $t$ from state $s_i$ to $s_j$, given the observations $Y$ (note that, for ease of notation, we drop the subscript and superscript of $Y_1^T$ when denoting the set of all observations, i.e., $Y=Y_1^T=\{y_1,y_2,...,y_T\}$) and the model as
\begin{align}
\epsilon_t(i,j) &=P(z_t=s_i,z_{t+1}=s_j|Y,\lambda)\nonumber \\
                &=\frac{\alpha_t(i)A_{ij}B_{jy_{t+1}}\beta_{t+1}(j)}{\sum_{k=1}^{N_s}{\sum_{l=1}^{N_s}} \alpha_t(k)A_{kl}B_{ly_{t+1}}\beta_{t+1}(l)}.&
\end{align}
 Here, if we sum $\epsilon_t(i,j)$ over the time index $t$, then we obtain the expected number of transitions from state $s_i$ to $s_j$ in the hidden state sequence $Z=\{z_1,z_2,...,z_{T}\}$. Next, we define the probability of being at state $z_t=s_i$, given the observations and the model as
\begin{align}
\gamma_t(i)=\sum_{j=1}^{N_s}\epsilon_t(i,j). \label{eq:gamaref1}
 \end{align}
Similarly, the summation of $\gamma_t(i)$ over the time index $t$ yields the expected number of times in the state $s_i$.

Then, given the definitions in \eqref{eq:recalpha}-\eqref{eq:gamaref1} and the observation sequence $Y$, we estimate the HMM parameters using the likelihood maximization through the EM algorithm \cite{EMalgorithm}, i.e., $\hat{\lambda}=\arg\max_{\lambda} P(Y|\lambda).$ The iterative EM equations that solve this maximization problem (at least locally) are as follows:
\begin{align}
\hat{A}_{ij} = \frac{\sum_{t=1}^{T-1}\epsilon_t(i,j)}{\sum_{t=1}^{T-1}\gamma_t(i)}, \text{  } \hat{B}_{ij} = \frac{\sum_{t=1}^{T-1} 1_{\{y_t=v_j\}}\gamma_t(i)}{\sum_{t=1}^{T-1} \gamma_t(i)}, \text{  } \hat{\pi}_i=\gamma_1(i). \label{eq:reestimation}
\end{align}
\noindent
Here, given the training data, we estimate the HMM parameters $\lambda$ by the iterative re-estimation procedure defined by the EM algorithm. Namely, given the HMM parameters $\lambda_{q-1}$ at an iteration $q$, we re-estimate the model parameters as $\lambda_{q}$ using the re-estimation formulas in \eqref{eq:reestimation}. This procedure is guaranteed to improve the likelihood of the observations at every iteration and converge to a set of HMM parameters $\hat{\lambda}$, which is at least locally optimal (cf. \cite{HMM} and the references therein).

In the following section, we derive the new set of iterative EM equations that incorporates the noisy side information into the HMM framework.

\section{HMM training with noisy and partial access to the state sequence \label{sec:deterministic_analysis}}
In this section, we derive the new set of iterative EM equations for the HMM parameter learning, when we have noisy side information on the hidden states. Here, we have an observation sequence
$y_t \in Y=\{y_1,y_2,...,y_T\}$, with \emph{partial} and \emph{noisy} access to the hidden states, $z_t \in Z = \{z_1,z_2,...,z_T\}$, as this side information. Each hidden state $z\in Z$ might be observed as $x$ with probability $\tau$, i.e., we do not necessarily have a state observation at a given time instant. Hence, we have \emph{partial} access to the hidden state sequence. In addition to this partial access, a state observation $x$ might also be \emph{noisy} such that $P(z \neq x)= (1-p)$.
\noindent
We assume that if an error happens, then $P(x = s) = \frac{1}{N_s-1},$  $\forall s \in S$  s.t. $z \neq s$. For ease of notation, we define the state observations at every time $t$ as $x_t\in X=\{x_1,x_2,...,x_T\}$, such that if $z_t$ is ever observed as $x$, then $x_t=x$. Otherwise, $x_t=s_0$, where $s_0$ is a pseudo-state. This expands our state space to $S'=S\cup\{s_0\}$. Thus, we model mislabeling and partial labeling jointly in one complete framework as shown in Fig. \ref{HMMfigure}b.

In order to incorporate the side information into the new framework, we first define the updated variables of the forward-backward procedure, which will be later used in derivation of the new set of iterative EM equations. The updated forward variable,
\begin{align}
\bar{\alpha}_{t}(i)=P(Y_{1}^{t},X_{1}^{t},z_t=s_i|\lambda),& \label{eq:alpha1}
\end{align}
is the probability of observing $(Y_{1}^{t},X_{1}^{t}=       \{x_1,x_2,...,x_t\}      )$ and being at state $z_t=s_i$, given the model $\lambda$. Note that $z_t$ is the correct and the underlying hidden state, whereas $X_{1}^{t}$ are the state observations, for which we might have: (1) $x_t=s_0$ corresponding to the case that $z_t$ is not actually observed and (2) noisy, if $z_t$ is actually observed. Similarly, the backward variable,
\begin{align}
\bar{\beta}_{t}(i)& =P(Y_{t+1}^{T},X_{t+1}^{T}|z_t=s_i, \lambda), \label{eq:bbeta1}
\end{align}
is the probability of observing $(Y_{t+1}^{T},X_{t+1}^{T}=\{x_{t+1},x_{t+2},...,x_T\})$, given the model and the state $z_t=s_i$. The updated forward and backward variables are the key variables of the new framework, which incorporate the side information. The following proposition explicitly relates these variables to the side information and provides the corresponding recursions.\\\\
{\bf Proposition 1:} For the updated forward and backward variables defined in \eqref{eq:alpha1} and \eqref{eq:bbeta1}, we have
\begin{align*}
\bar{\alpha}_{t}(i)& =\nu(x_t,s_i)B_{iy_t}\sum_{j=1}^{N_s}A_{ji}\bar{\alpha}_{t-1}(j),
\text{     $2\leq t \leq T$},\\
\bar{\beta}_{t}(i)& = \sum_{j=1}^{N_s} \nu(x_{t+1},s_j) \bar{\beta}_{t+1}(j)A_{ij} B_{jy_{t+1}},
\text{$1\leq t \leq T-1$},
\end{align*}
where $\nu(x_t,s_i)=1_{\{ x_t=s_0 \}}(1-\tau) +     1_{\{x_t=s_i\}}\tau p + 1_{\{x_t\neq s_i \wedge x_t \neq s_0 \}}\frac{\tau(1-p)}{N_s-1}$, $s_i \neq s_0$, $s_j \neq s_0$ and $1_{ \{h\} }$ is the indicator function such that  $1_{ \{h\} }=1$ if $h$ is true and $1_{ \{h\} }=0$, otherwise.\\\\
{\bf Proof:} Using the marginalization over the random variable $z_{t-1}$, we can obtain $\bar{\alpha}_{t}(i)$ as
\begin{align*}
               \bar{\alpha}_{t}(i)&=\sum_{j=1}^{N_s}P(Y_{1}^{t},X_{1}^{t},z_t=s_i,z_{t-1}=s_j|\lambda),
\end{align*}
which can be expressed, using the product of conditional probabilities, as
\begin{align*}
               \bar{\alpha}_{t}(i)&=\sum_{j=1}^{N_s}P(y_t,x_t,z_t=s_i|Y_{1}^{t-1},X_{1}^{t-1},z_{t-1}=s_j,\lambda)P(Y_{1}^{t-1},X_{1}^{t-1},z_{t-1}=s_j|\lambda).
\end{align*}
By definition of the updated forward variable, we get
\begin{align*}
               \bar{\alpha}_{t}(i)&=\sum_{j=1}^{N_s}P(y_t,x_t,z_t=s_i|Y_{1}^{t-1},X_{1}^{t-1},z_{t-1}=s_j,\lambda)\bar{\alpha}_{t-1}(j),
\end{align*}
where Markov Property is applied to reach
\begin{align*}
               \bar{\alpha}_{t}(i)&=\sum_{j=1}^{N_s}P(y_t,x_t,z_t=s_i|z_{t-1}=s_j,\lambda)\bar{\alpha}_{t-1}(j)\\
               &=\sum_{j=1}^{N_s}P(y_t,x_t|z_t=s_i,z_{t-1}=s_j,\lambda)P(z_t=s_i|z_{t-1}=s_j,\lambda)\bar{\alpha}_{t-1}(j)\\
               &=\sum_{j=1}^{N_s}P(y_t,x_t|z_t=s_i,\lambda)P(z_t=s_i|z_{t-1}=s_j,\lambda)\bar{\alpha}_{t-1}(j).
\end{align*}
Since $x_t$ and $y_t$ are independent conditioned on ($z_t, \lambda$), we obtain
\begin{align*}
               \bar{\alpha}_{t}(i)&=\sum_{j=1}^{N_s}P(y_t,x_t|z_t=s_i,\lambda)A_{ji}\bar{\alpha}_{t-1}(j)\\
               &=\sum_{j=1}^{N_s}P(x_t|z_t=s_i,\lambda)P(y_t|z_t=s_i,\lambda)A_{ji}\bar{\alpha}_{t-1}(j).
\end{align*}
Then, by definition of the probability of error events in the side information, we get the proposition for the updated forward variable as
\begin{align*}
               \bar{\alpha}_{t}(i)&=\nu(x_t,s_i)B_{iy_t}\sum_{j=1}^{N_s}A_{ji}\bar{\alpha}_{t-1}(j), \text{  }2\leq t\leq T.
\end{align*}
As for the initialization, we set $\bar{\alpha}_{1}(i)=\nu(x_1,s_i)\pi_iB_{iy_1}$. Similarly, the corresponding recursion for the updated backward variable can be found as
\begin{align*}
               \bar{\beta}_{t}(i)& = \sum_{j=1}^{N_s} \nu(x_{t+1},s_j) \bar{\beta}_{t+1}(j)A_{ij} B_{jy_{t+1}}, \text{  }1\leq t\leq T-1,
\end{align*}
for which we have the initialization $\bar{\beta}_T(i)=1$.$\blacksquare$\\
Here, $p$ reflects the confidence that we have on the side information and it is a parameter in our PHMM training algorithm. Ideally, when given a set of data, $p$ (named as $p_{\mathrm{train}}$ in Section IV) should be set according to the underlying true noise level, $1-p_{\mathrm{true}}$, which is unknown. This brings an immediate trade-off between setting the confidence too low or too high, when an accurate guess about $1-p_{\mathrm{true}}$ is not present. If we have too high confidence, then our algorithm basically overfits to the noise in the side information, which degrades the state recognition performance as discussed in Section IV. On the other hand, if we have too low confidence, then our algorithm does not fully exploit the side information to its limit. We discuss this later in Section IV, when investigating the robustness of our algorithm to the confidence parameter $p$ ($p_{\mathrm{train}}$ in Section IV).

We next define the probability of transition at time $t$ from state $s_{i}$ to $s_{j}$, given the observations $Y$, the side information $X$, and the model as
\begin{align}
\bar{\epsilon}_t(i,j)=P(z_t=s_{i},z_{t+1}=s_{j}|Y,X,\lambda), \label{eq:epsilon1}
\end{align}
which is essential to the estimation of the HMM parameters in our new framework. Note that the summation of $\bar{\epsilon}_t(i,j)$ over the time index $t$ is the expected number of transitions from state $s_i$ to $s_j$, when we have side information in addition to the observation sequence.

The following proposition relates $\bar{\epsilon}_t(i,j)$ to the updated forward and backward variables.\\\\
\noindent
{\bf Proposition 2:} With the definitions in \eqref{eq:alpha1} and \eqref{eq:bbeta1}, we have
\begin{align*}
\bar{\epsilon}_t(i,j)&=P(z_t=s_{i},z_{t+1}=s_{j}|Y,X,\lambda) \\
                           &=\frac{B_{jy_{t+1}}\nu(x_{t+1},s_j) A_{ij}\bar{\alpha}_{t}(i)\bar{\beta}_{t+1}(j)}{P(Y,X|\lambda)},
\end{align*}
where $\nu(x_{t+1},s_j)=1_{\{ x_{t+1}=s_0 \}}(1-\tau) +     1_{\{x_{t+1}=s_j\}}\tau p + 1_{\{x_{t+1}\neq s_j \wedge x_{t+1} \neq s_0 \}}\frac{\tau(1-p)}{N_s-1}$, $s_j \neq s_0$ and $1_{ \{h\} }$ is the indicator function such that  $1_{ \{h\} }=1$ if $h$ is true and $1_{ \{h\} }=0$, otherwise.\\
\\
\noindent
{\bf Proof:}
Splitting the observations as $Y=(Y_1^t,y_{t+1},Y_{t+2}^T)$, and the side information as $X=(X_1^t,x_{t+1},X_{t+2}^T)$, \eqref{eq:epsilon1} yields
\begin{align*}
\bar{\epsilon}_t(i,j)&=\frac{P(z_t=s_{i},z_{t+1}=s_{j},X_1^t,x_{t+1},X_{t+2}^T,Y_1^t,y_{t+1},Y_{t+2}^T|\lambda)}{P(Y,X|\lambda)}\nonumber\\
& =\frac{P(z_t=s_{i},X_{1}^t,x_{t+1},Y_1^t,y_{t+1}|z_{t+1}=s_{j},X_{t+2}^T,Y_{t+2}^T,\lambda)P(z_{t+1}=s_{j},X_{t+2}^T,Y_{t+2}^T|\lambda)}{P(Y,X|\lambda)}.\nonumber
\end{align*}
Since $(z_t=s_{i},X_{1}^t,x_{t+1},Y_1^t,y_{t+1})$ is independent with $(X_{t+2}^T, Y_{t+2}^T)$ conditioned on $(z_{t+1},\lambda)$, we obtain
\begin{align*}
\bar{\epsilon}_t(i,j)&=\frac{P(z_t=s_{i},X_{1}^t,x_{t+1},Y_1^t,y_{t+1}|z_{t+1}=s_{j},\lambda)P(z_{t+1}=s_{j},X_{t+2}^T,Y_{t+2}^T|\lambda)}{P(Y,X|\lambda)},\nonumber
\end{align*}
which, re-arranging the conditional probabilities, yields
\begin{align*}
\bar{\epsilon}_t(i,j) &=\frac{P(z_t=s_{i},z_{t+1}=s_{j},X_{1}^t,x_{t+1},Y_1^t,y_{t+1}|\lambda)P(X_{t+2}^T,Y_{t+2}^T|z_{t+1}=s_{j},\lambda)}{P(Y,X|\lambda)} \nonumber\\ &=\frac{P(z_{t+1}=s_{j},x_{t+1},y_{t+1}|z_t=s_{i},X_1^{t},Y_1^{t},\lambda)P(z_t=s_{i},X_1^{t},Y_1^{t}|\lambda)P(X_{t+2}^T,Y_{t+2}^T|z_{t+1}=s_{j},\lambda)}{P(Y,X|\lambda)}. \nonumber
\end{align*}
Since $(z_{t+1}=s_{j},x_{t+1},y_{t+1})$ and $(X_1^{t},Y_1^{t})$ are independent conditioned on $(z_t=s_{i},\lambda)$, and recognizing the terms $\bar{\alpha}_{t}(i)$ and $\bar{\beta}_{t+1}(j)$, we obtain
\begin{align*}
\bar{\epsilon}_t(i,j) &=\frac{P(z_{t+1}=s_{j},x_{t+1},y_{t+1}|z_t=s_{i},\lambda)\bar{\alpha}_{t}(i)\bar{\beta}_{t+1}(j)}{P(Y,X|\lambda)}\\ \nonumber
&=\frac{P(x_{t+1},y_{t+1}|z_{t+1}=s_{j},z_t=s_{i},\lambda)P(z_{t+1}=s_{j}|z_t=s_{i},\lambda)\bar{\alpha}_{t}(i)\bar{\beta}_{t+1}(j)}{P(Y,X|\lambda)}, \nonumber
\end{align*}
wherein, Markov Property is used to reach
\begin{align*}
\bar{\epsilon}_t(i,j) &=\frac{P(x_{t+1},y_{t+1}|z_{t+1}=s_{j},\lambda)P(z_{t+1}=s_{j}|z_t=s_{i},\lambda)\bar{\alpha}_{t}(i)\bar{\beta}_{t+1}(j)}{P(Y,X|\lambda)}. \nonumber
\end{align*}
Since $x_{t+1}$ is independent with $y_{t+1}$ conditioned on $(z_{t+1},\lambda)$, we obtain
\begin{align*}
\bar{\epsilon}_t(i,j) &=\frac{P(y_{t+1}|z_{t+1}=s_{j},\lambda)P(x_{t+1}|z_{t+1}=s_{j},\lambda)P(z_{t+1}=s_{j}|z_t=s_{i},\lambda)\bar{\alpha}_{t}(i)\bar{\beta}_{t+1}(j)}{P(Y,X|\lambda)}. \nonumber
\end{align*}
Then, due to the definition of the probability of error event in the side information, we get the proposition as
\begin{align*}
&\bar{\epsilon}_t(i,j) =\frac{B_{jy_{t+1}}\nu(x_{t+1},s_j) A_{ij}\bar{\alpha}_{t}(i)\bar{\beta}_{t+1}(j)}{P(Y,X|\lambda)}.\blacksquare
\end{align*}
\noindent
Finally, we define the probability of being at state $z_t=s_{i}$, given the observations, the side information and the model as
\begin{align}
\bar{\gamma}_t(i)=P(z_t=s_{i}|Y,X,\lambda)=\sum_{j=1}^{N_s}\bar{\epsilon}_t(i,j). \label{eq:gamma}
\end{align}
Note that summation of $\bar{\gamma}_t(i)$ over the time index $t$ is the expected number of times we are in the state $s_i$, when we have side information in addition to the observation sequence.

Next, we derive the new set of equations that incorporates the side information $X$. For this, the parameter set $\hat{\lambda}$ will be selected such that the log-likelihood of the training data (observations and the side information), i.e., $F = \log\big(P(X,Y|\lambda)\big)$, is maximized via using an auxiliary function \cite{HMM,aux,FBP2}. Instead of maximizing $F$, we maximize the auxiliary function $F'$ through the EM algorithm.
Let $Q(Z)=P(Z|X,Y,\lambda')$ be the output of E-step. Then, with respect to $\lambda$, M-step maximizes
\begin{align*}
F'=\sum_{Z}Q(Z)\log(\frac{P(X,Y,Z|\lambda)}{Q(Z)}).
\end{align*}
\noindent
The following theorem provides the main result of our work for incorporating the side information within the framework of the basic HMM parameter learning problem.\\\\
{\bf Theorem:} With the definitions in \eqref{eq:epsilon1} and \eqref{eq:gamma}, the maximization of $F'$ through the EM algorithm is convergent (at least locally) to
\begin{align*}
\hat{A}_{ij} = \frac{\sum_{t=1}^{T-1}\bar{\epsilon}_t(i,j)}{\sum_{t=1}^{T-1}\bar{\gamma}_t(i)},
\hat{B}_{ij} = \frac{\sum_{t=1}^{T-1}   1_{\{y_t=v_j\}} \bar{\gamma}_t(i)}
{\sum_{t=1}^{T-1} \bar{\gamma}_t(i)}.
\end{align*}\\
{\bf Proof:}
We give an outline for the proof. Let $Q(Z)=P(Z|X,Y,\lambda')$ be the output of E-step, then M-step carries out the following maximization:
\begin{align}
\arg\max_\lambda F' &=\arg\max_\lambda \sum_{Z}Q(Z)\log(\frac{P(X,Y,Z|\lambda)}{Q(Z)}) \label{eq:LogmaximizationBeforeSplitting}\\
                    &=\arg\max_\lambda \sum_{Z}Q(Z)\bigg(\log\big(P(X,Y,Z|\lambda)\big)-\log\big(Q(Z)\big)\bigg). \label{eq:LogmaximizationAfterSplitting}
\end{align}
Since $Q(Z)$ is the output of E-step, it does not depend on $\lambda$. Hence, if we split the log division in \eqref{eq:LogmaximizationBeforeSplitting} into subtraction, then we can drop the second term (subtrahend) in \eqref{eq:LogmaximizationAfterSplitting} and obtain the maximization
\begin{align}
\arg\max_\lambda F' &= \arg\max_\lambda \sum_{Z}Q(Z)\log\big(P(X,Y,Z|\lambda)\big),\nonumber
\end{align}
which, using the product of conditional probabilities, yields
\begin{align}
\arg\max_\lambda F' &=\arg\max_\lambda \sum_{Z}Q(Z)\log\big(P(Y|X,Z,\lambda) P(X|Z,\lambda) P(Z|\lambda)\big).\nonumber
\end{align}
Since $X$ is independent with $\lambda$ conditioned on $Z$ and $Y$ is independent with $X$ conditioned on $(Z,\lambda)$, we obtain
\begin{align}
\arg\max_\lambda F' &=\arg\max_\lambda \sum_{Z}Q(Z)\log\big(P(Y|Z,\lambda) P(X|Z) P(Z|\lambda)\big),\nonumber
\end{align}
where we can drop the term $P(X|Z)$ since it does not depend on $\lambda$ and reach
\begin{align}
\arg\max_\lambda F' &=\arg\max_\lambda \sum_{Z}Q(Z)\log\big(P(Y|Z,\lambda) P(Z|\lambda)\big).\label{eq:theorem4}
\end{align}
We point out that the maximization in \eqref{eq:theorem4} does not involve the side information $X$, except that $Q(Z)=P(Z|X,Y,\lambda')$ is related to $X$. However, since $Q(Z)$ is calculated in E-step before M-step starts in the course of our algorithm, it only brings constant factors to the maximization in \eqref{eq:theorem4} and, hence, it does not affect the M-step derivations. Therefore, rest of the derivations follows the regular M-step derivations of the EM algorithm for the ordinary HMM parameter training and we estimate the transition probabilities as
\begin{align*}
\hat{A}_{ij} &=\frac{\sum_{Z}Q(Z)\sum_{t=1}^{T-1} 1_{\{z_{t}=s_i \wedge z_{t+1} = s_j\}}}{\sum_{Z}Q(Z)\sum_{t=1}^{T-1} 1_{\{z_{t}=s_i\}}}\\
&=\frac{\sum_{Z}\sum_{t=1}^{T-1} 1_{\{z_{t}=s_i \wedge z_{t+1} = s_j\}} P(Z|X,Y,\lambda')}{\sum_{Z}\sum_{t=1}^{T-1} 1_{\{z_{t}=s_i\}}P(Z|X,Y,\lambda')},
\end{align*}
where the indicator function in the numerator and the denominator marginalizes the probability $P(Z|X,Y,\lambda')$, since the outer summation is over all possible hidden state sequences. Hence, we obtain
\begin{align*}
\hat{A}_{ij} &=\frac{\sum_{t=1}^{T-1} P(z_{t}=s_i,z_{t+1}=s_j|X,Y,\lambda')}{\sum_{t=1}^{T-1} P( z_{t}=s_i|X,Y,\lambda')}\\
             &=\frac{\sum_{t=1}^{T-1}\bar{\epsilon}_t(i,j)}{\sum_{t=1}^{T-1}\bar{\gamma}_t(i)},
\end{align*}
which is, given the side information, the expected number of transitions from state $s_i$ to $s_j$ divided by the expected number of times in the state $s_i$. Similarly, $\hat{B}_{ij}$ is given by
\begin{align*}
\hat{B}_{ij} = \frac{\sum_{t=1}^{T-1}   1_{\{y_t=v_j\}} \bar{\gamma}_t(i)}
{\sum_{t=1}^{T-1} \bar{\gamma}_t(i)},
\end{align*}
which is, given the side information, the expected number of times in the state $s_i$ and observing $v_j$, divided by the expected number of times in the state $s_i$.
Also, the set of initial probabilities for the hidden state $z_1$ is estimated as $\hat{\pi}_i = \bar{\gamma}_1(i). \blacksquare$

Based on the new set of equations as well as the recursions defined in Proposition 1,
we incorporated possibly corrupted side information into the HMM training framework. In the next section, we provide examples that demonstrate the performance of the new set of training updates under different scenarios.
\section{Simulations}
In this section, we demonstrate the performance of our method through simulations using data generated with the following HMM parameters:\\

\noindent
{\small
$N_s=3$, $N_v=3$, $ \pi = \begin{bmatrix} 0.3 & 0.3 & 0.4 \end{bmatrix}$, and
$A=\left[\begin{array}{ccc} 0.8 & 0.19 & 0.01 \\ 0.01 & 0.8 & 0.19 \\ 0.19 & 0.01 & 0.8 \\ \end{array} \right]$, $ B = \begin{bmatrix} 0.6 & 0.3 & 0.1 \\ 0.1 & 0.6 & 0.3 \\ 0.3 & 0.1 & 0.6 \\ \end{bmatrix}$.}

\noindent
For these simulations, we have a test set of 500 data points and a training set of 250 data points along with the side information of a relatively high noise level, $1-p_{\mathrm{true}}=0.4$, and a relatively low noise level, $1-p_{\mathrm{true}}=0.2$, with $\tau$ ranging from 0 to 0.6. We emphasize that the exact noise level may not be known by the algorithm. Hence, we provide $p_{\mathrm{train}}$ to the algorithm which may not be equal to the $p_{\mathrm{true}}$. Here, the parameter $p_{\mathrm{train}}$ reflects the confidence (equivalently the expected noise level) that we have on the side information. Since this confidence on the side information might not be accurate, i.e., $p_{\mathrm{train}}$ does not necessarily match with $p_{\mathrm{true}}$, for analyzing the sensitivity of our method to the confidence parameter, we train our algorithm with different choices for $p_{\mathrm{train}}$: (1) we set confidence that is in the proximity of $p_{\mathrm{true}}$ ($p_{\mathrm{train}}\sim p_{\mathrm{true}}$), i.e., $p_{\mathrm{train}}\in\{0.55,0.6,0.65\}$, if $p_{\mathrm{true}}=0.6$ and $p_{\mathrm{train}}\in\{0.75,0.8,0.85\}$, if $p_{\mathrm{true}}=0.8$, (2) we set too high confidence on the side information ($p_{\mathrm{train}}\gg p_{\mathrm{true}}$), i.e.,  $p_{\mathrm{train}}=1$, when $p_{\mathrm{true}}\in \{0.6,0.8\}$ and, (3) we set too low confidence ($p_{\mathrm{train}}\ll p_{\mathrm{true}}$), i.e., $p_{\mathrm{train}}=0.5$, when $p_{\mathrm{true}}=1$. Using the training set, we first estimate the unknown model parameters, $A_{ij}$, $B_{ij}$, and $\pi_{ij}$. Then, on the test set, the hidden state sequence is estimated by the Viterbi algorithm \cite{Viterbi1, Viterbi2} using the estimated model parameters. This process is repeated 500 times and we present the average state recognition error rates for all the cases aforementioned. In order to show the efficacy of incorporating the side information by our method, we compare the state recognition error rates of our algorithm with: (1) Baseline Performance, the state recognition error rate if the model parameters are estimated by the ordinary HMM parameter estimation. This is the performance, which is readily achievable with no side information. (2) The Oracle, the state recognition error rate if the true model parameters are directly used in the state estimation on the test set. This is the performance limit if the HMM training algorithm is run on infinite amount of training data, which is only asymptotically achievable. (3) Limit of Algorithm, the state recognition error rate if the side information is completely accurate and the algorithm is trained with complete confidence on the side information, i.e., $p_{\mathrm{true}}=1$, $p_{\mathrm{train}}=1$. Finally, this is the performance limit that our algorithm can gain at most by exploiting the side information. Here, we name the difference between the Baseline Performance and the Oracle as the ``achievable margin" since no algorithm can obtain state recognition improvements more than the achievable margin, provided that, as in this work, first the model parameters are estimated and then used in the Viterbi algorithm for state recognition.
\begin{figure}[t]
\centerline{\epsfxsize=24.5cm \epsfysize=10.5cm  \epsfbox{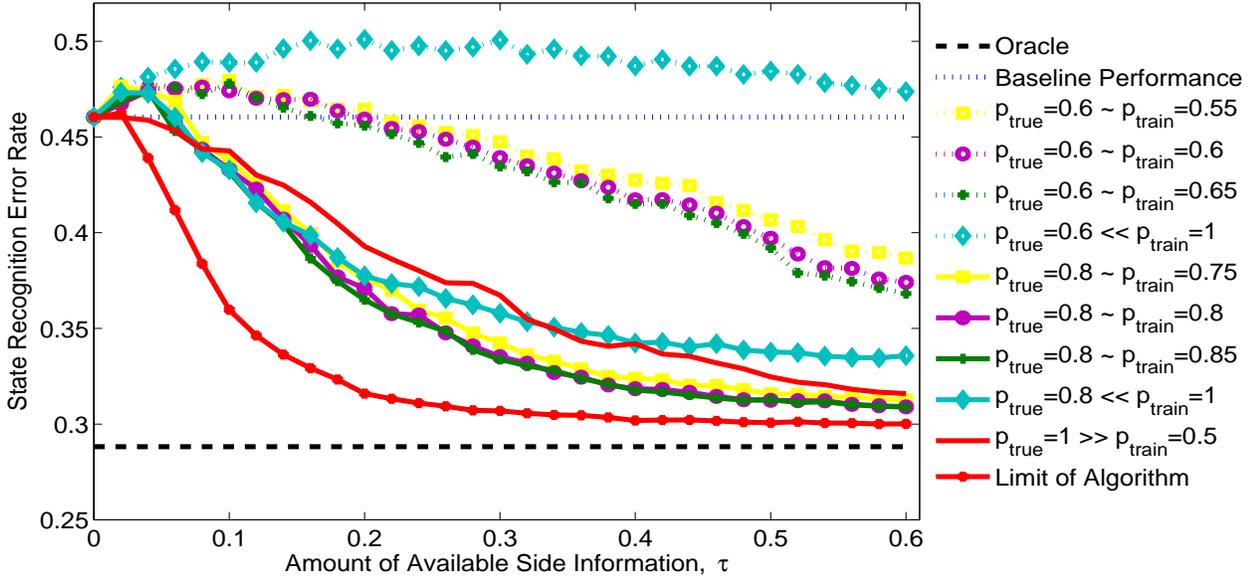}}
\caption{{\small Simulation results for different scenarios. Our algorithm is trained with $p_{\mathrm{train}} \in\{0.55,0.60,0.65\}$ when $p_{\mathrm{true}}=0.60$ and $p_{\mathrm{train}} \in\{0.75,0.80,0.85\}$ when $p_{\mathrm{true}}=0.80$. The State Recognition Error Rates are estimated by the Viterbi algorithm. Performance of our algorithm is compared against three performance limits: (1) Baseline Performance, error rate by ordinary HMM using no side Information, (2) Oracle, error rate if the true model parameters are used in state recognition, and (3) Limit of Algorithm, $p_{\mathrm{train}}=p_{\mathrm{true}}=1$. See the text for details.}}
\label{Performance}
\end{figure}

Our simulations show that the performance of our method, provided that $p_{\mathrm{train}}\sim p_{\mathrm{true}}$, improves with the amount of side information that is indicated by $\tau$. In particular, when we have accurate access to the hidden states, i.e., $p_{\mathrm{true}}=p_{\mathrm{train}}=1$, the state recognition rate in the test set, labeled as Limit of Algorithm in Fig. \ref{Performance}, consistently approaches to the Oracle as $\tau$ increases and reaches $\sim 90\%$ gain (the performance improvement over the baseline corresponds to $\sim 90\%$ of the achievable margin) with $30\%$ additional information on states, i.e., $\tau=0.3$, as shown in Fig. \ref{Performance}. This proves the efficacy of our method with incorporating the side information. On the other hand, in the case of noisy access to the hidden states such that $20\%$ of the state observations are mislabeled, i.e., $p_{\mathrm{true}}=0.8$, our method (when $p_{\mathrm{train}}\sim p_{\mathrm{true}}$) is able to provide substantial gain, $70\%$, at $\tau=0.3$. In this case, as $\tau$ increases, the recognition approaches to Limit of Algorithm showing that our algorithm optimally incorporates the side information under noise asymptotically. Even if the noise level is further increased up to a level as high as $40 \%$ mislabeling, we still obtain a gain that consistently increases with $\tau$, when $p_{\mathrm{train}}\sim p_{\mathrm{true}}$. Thus, our method is robust to noise. Nevertheless, the algorithm must not rely on the side information with too high confidence. Specifically, when we have the confidence $p_{\mathrm{train}}=1$ in case of high noise level, i.e., $p_{\mathrm{true}}=0.6$, we do not obtain any improvement compared to the baseline. On the contrary, the algorithm does not fully exploit the side information to its limit, if the confidence is too low. For instance, in case of $p_{\mathrm{train}}=0.5$ and $p_{\mathrm{true}}=1$, the rate of performance improvement with $\tau$ is significantly slower than that of Limit of Algorithm, i.e., $p_{\mathrm{true}}=1$, $p_{\mathrm{train}}=1$. According to our simulations, setting the confidence in the proximity of the true noise level is sufficient to obtain the maximum gain, i.e., our algorithm does not require an exact match between $p_{\mathrm{true}}$ and $p_{\mathrm{train}}$. This demonstrates that our algorithm is also robust to the mismatches in the confidence parameter $p_{\mathrm{train}}$.
\label{sec:simulations}
\section{Conclusion}
\label{sec:conclusion}
In this paper, we introduced a new parameter estimation algorithm for HMM, when we have partial and noisy access to the hidden state sequence as side information. This side information can be seen as partial labeling, ``possibly wrong", of the hidden states. In this work, we model mislabeling and partial labeling of the hidden states jointly in one complete framework. This framework naturally recovers the
unsupervised HMM training if the partial access to the hidden states
is turned off. In our simulations, we observed that, using this side information, we considerably improved the state recognition performance, up to $70\%$, with respect to the ``achievable margin". Moreover, our method is shown to be robust to the training conditions. Finally, since this framework includes possible mislabeling events, our algorithm models realistic training conditions more accurately than the ordinary HMM training. Hence, we expect the same performance improvement in other examples.
\section{Acknowledgements}
This work was partially supported by Turk Telekom under Grant Number 11315-05.

\bibliographystyle{elsarticle-num}
\bibliography{msaf_references}

\begin{thebibliography}{10}
\expandafter\ifx\csname url\endcsname\relax
  \def\url#1{\texttt{#1}}\fi
\expandafter\ifx\csname urlprefix\endcsname\relax\def\urlprefix{URL }\fi
\expandafter\ifx\csname href\endcsname\relax
  \def\href#1#2{#2} \def\path#1{#1}\fi

\bibitem{HMM}
L.~R. Rabiner, A tutorial on hidden markov models and selected applications in
  speech recognition, Proc. IEEE 77 (1989) 257--286.

\bibitem{DREFH}
K.~K. Paliwal, Dimensionality reduction of the enhanced feature set for the
  hmm-based speech recognizer, Digital Signal Processing 2 (1992) 157 -- 173.

\bibitem{SRUGHSMM}
M.~D. Moore, M.~I. Savic, Speech reconstruction using a generalized hsmm
  (ghsmm), Digital Signal Processing 14 (2004) 37 -- 53.

\bibitem{APIOAHMMWDMFSR}
C.~Mitchell, M.~Harper, L.~Jamieson, R.~Helzerman, A parallel implementation of
  a hidden markov model with duration modeling for speech recognition, Digital
  Signal Processing 5 (1995) 43 -- 57.

\bibitem{APPT}
D.~Cutting, J.~Kuipec, J.~Pedersen, P.~Sibun, A practical part-of-speech
  tagger, Proc. Third Conf. Applied Natural Language Processing (1992) 133 --
  140.

\bibitem{LSDTHMMFSR}
P.~C. Woodland, D.~Povey, Large scale discriminative training of hidden markov
  models for speech recognition, Computer Speech and Language 16 (2002) 25 --
  47.

\bibitem{ELLAFSR}
S.~Kozat, K.~Visweswariah, R.~Gopinath, Efficient, low latency adaptation for
  speech recognition, IEEE Int. Conf. on Acoustic Speech and Signal Processing
  (2007) 777 -- 780.

\bibitem{HMMIBSA}
E.~Birney, Hidden markov models in biological sequence analysis, IBM journal of
  Research and Development 45 (2001) 449.

\bibitem{HMMbi}
V.~Fonzo, F.~Aluffi-Pentini, V.~Parisi, Hidden markov models in bioinformatics,
  in: Current Bioinformatics, 2007, pp. 49 -- 61.

\bibitem{SIFPHMM}
L.~Bordes, P.~Vandekerkhove, Statistical inference for partially hidden markov
  models, Commun. in Statistics 34 (2005) 1081--1104.

\bibitem{TETWAPM}
B.~Merialdo, Tagging english text with a probabilistic model, Comput. Linguist.
  20 (1994) 155--171.

\bibitem{DBWRHT}
D.~Elworthy, Does baum-welch re-estimation help taggers?, in: Proceedings of
  the fourth conference on Applied natural language processing, ANLC '94, 1994,
  pp. 53--58.

\bibitem{LHMMSFIE}
K.~Seymore, A.~Mccallum, R.~Rosenfeld, Learning hidden markov model structure
  for information extraction, In AAAI 99 Workshop on Machine Learning for
  Information Extraction (1999) 37--42.

\bibitem{ALOPHMM}
T.~Scheffer, S.~Wrobel, Active learning of partially hidden markov models,
  Proc. of ECML/PKDD Workshop on Instance Selection.

\bibitem{EMalgorithm}
A.~P. Dempster, N.~M. Laird, D.~B. Rubin, Maximum likelihood from incomplete
  data via the em algorithm, Journal of the Royal Statistical Society 39 (1977)
  1 -- 38.

\bibitem{BaumWelch}
L.~E. Baum, T.~Petrie, G.~Soules, N.~Weiss, A maximization technique occurring
  in the statistical analysis of probabilistic functions of markov chains, The
  Annals of Mathematical Statistics 41 (1970) 164--171.

\bibitem{PHMM}
S.~Forchhammer, J.~Rissanen, Partially hidden markov models, IEEE Transactions
  on Information Theory 42 (1996) 1253--1256.

\bibitem{SSL}
O.~Chapelle, B.~Schölkopf, A.~Zien, Semi-supervised learning (adaptive
  computation and machine learning), MIT Press.

\bibitem{FBP1}
L.~E. Baum, J.~A. Eagon, An inequality with applications to statistical
  estimation for probabilistic functions of markov processes and to a model for
  ecology, Bulletin of the American Mathematical Society 73 (1967) 360 -- 363.

\bibitem{FBP2}
L.~E. Baum, G.~R. Sell, Growth functions for transformations on manifolds,
  Pacific J. Math 27 (1968) 211 -- 227.

\bibitem{aux}
J.~Baker, The dragon system--an overview, in: Acoustics, Speech and Signal
  Processing, IEEE Transactions on, Vol.~23, 1975, pp. 24 -- 29.

\bibitem{Viterbi1}
A.~Viterbi, Error bounds for convolutional codes and an asymptotically optimum
  decoding algorithm, Information Theory, IEEE Transactions on 13 (1967) 260 --
  269.

\bibitem{Viterbi2}
G.~D. Forney, The viterbi algorithm, Proceedings of the IEEE 61 (1973) 268 --
  278.

\end{thebibliography}
\newpage
\end{document}